\documentclass[sigconf]{acmart}

\renewcommand\footnotetextcopyrightpermission[1]{} 
\usepackage{enumitem}
\usepackage{rotating, multirow}
\usepackage{fancyhdr}
\AtBeginDocument{%
  \providecommand\BibTeX{{%
    \normalfont B\kern-0.5em{\scshape i\kern-0.25em b}\kern-0.8em\TeX}}}

\setcopyright{acmcopyright}
\copyrightyear{2022}
\acmYear{2022}
\acmDOI{XXXXXXX.XXXXXXX}

\acmConference[MM '22]{the 30th ACM Int'l Conference on Multimedia}{October 10--14,
  2022}{Lisbon, Portugal}
\acmPrice{15.00}
\acmISBN{978-1-4503-XXXX-X/18/06}

\begin{document}
\title{ViT-FOD: A Vision Transformer based Fine-grained Object Discriminator}


\author{Zi-Chao Zhang}
\affiliation{
  \institution{School of Software, Shandong University}
  \city{Jinan}
  \country{China}}
\email{zhangzichao1008@163.com}

\author{Zhen-Duo Chen}
\authornote{Corresonding Author}
\affiliation{
  \institution{School of Software, Shandong University}
  \city{Jinan}
  \country{China}}
\email{chenzd.sdu@gmail.com}

\author{Yongxin Wang}
\affiliation{
  \institution{Shandong Jianzhu University}
  \city{Jinan}
  \country{China}}
\email{ yxinwang@hotmail.com}

\author{Xin Luo}
\affiliation{
  \institution{School of Software, Shandong University}
  \city{Jinan}
  \country{China}}
\email{luoxin.lxin@gmail.com}

\author{Xin-Shun Xu}
\affiliation{
  \institution{School of Software, Shandong University}
  \city{Jinan}
  \country{China}}
\email{xuxinshun@sdu.edu.cn}

\renewcommand{\shortauthors}{Zhang and Chen, et al.}

\begin{abstract}
Recently, several Vision Transformer (ViT) based methods have been proposed for Fine-Grained Visual Classification (FGVC).
These methods significantly surpass existing CNN-based ones, demonstrating the effectiveness of ViT in FGVC tasks.
However, there are some limitations when applying ViT directly to FGVC.
First, ViT needs to split images into patches and calculate the attention of every pair, which may result in heavy redundant calculation and unsatisfying performance when handling fine-grained images with complex background and small objects.
Second, a standard ViT only utilizes the class token in the final layer for classification, which is not enough to extract comprehensive fine-grained information.
To address these issues, we propose a novel ViT based fine-grained object discriminator for FGVC tasks, ViT-FOD for short.
Specifically, besides a ViT backbone, it further introduces three novel components, i.e, Attention Patch Combination (APC), Critical Regions Filter (CRF), and Complementary Tokens Integration (CTI).
Thereinto, APC pieces informative patches from two images to generate a new image so that the redundant calculation can be reduced.
CRF emphasizes tokens corresponding to discriminative regions to generate a new class token for subtle feature learning.
To extract comprehensive information, CTI integrates complementary information captured by class tokens in different ViT layers.
We conduct comprehensive experiments on widely used datasets and the results demonstrate that ViT-FOD is able to achieve state-of-the-art performance.
\end{abstract}




\maketitle

\section{Introduction}
Fine-Grained Visual Classification (FGVC) aims to recognize subordinate categories, such as bird categories\cite{cub, nabirds} and dog breeds\cite{dog}.
Due to the large intra-class variations and small inter-class variations, FGVC is much challenging. Most existing methods adopt a location-based feature extraction paradigm by focusing on subtle but discriminative parts.
Especially, with the development of deep CNNs\cite{resnet,vgg,densenet,googlenet}, significant progress has been made\cite{racnn, macnn, fdl}.
However, these CNN-based methods are gradually hitting a plateau.
One of the main reasons could be that the models based on CNN are naturally suitable for discovering discriminative regions spatially but lack appropriate means to establish relations between these regions and integrate them into a unified concept.
The self-attention mechanism is a solution to this problem.
Inspired by this, Vision Transformer (ViT)\cite{vit} with multiple self-attention layers has also been introduced into Computer Vision and attracted extensive attention.
More recently, several works\cite{transfg,ffvt,tpskg,rams} have tried to apply ViT to FGVC and make a breakthrough.

\begin{figure}
  \centering
  \includegraphics[width=.3\textwidth]{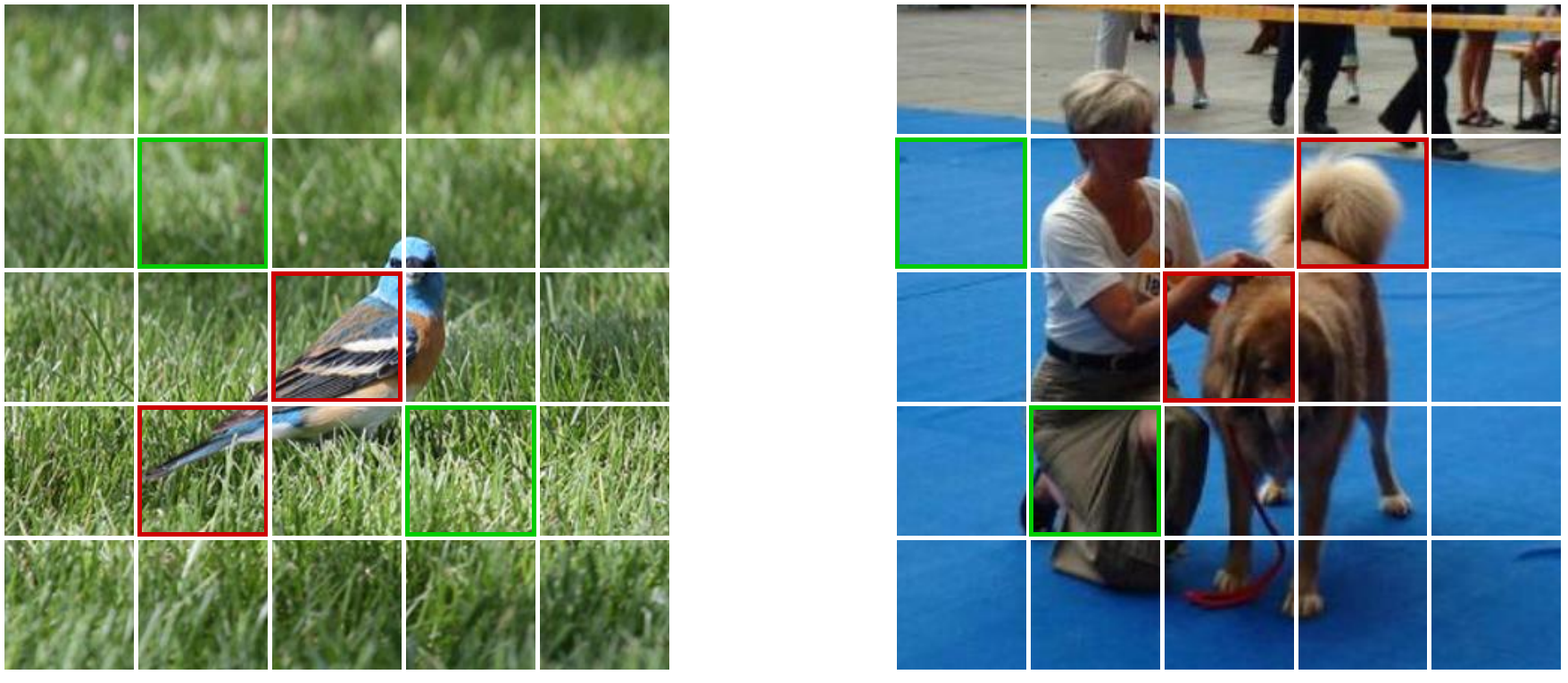}
  \caption{Example of redundancy calculation in ViT. The attention between red boxes is useful, and the calculation between green boxes is redundant.}
  \label{intro}
  \vspace{-0.3cm}
\end{figure}

These primary ViT-based attempts have significantly surpassed existing convolution-based methods, demonstrating the superiority of ViT in FGVC. However, there are still some issues to be further considered when applying ViT to this task. 1) A standard ViT model needs to split an image into patches as input, and then the Multi-head Self-Attention (MSA) module in each layer obtains the relationship between any two patches. However, for fine-grained images, many samples contain complex backgrounds, and some objects could be relatively small as well. When such images are processed by ViT, a large amount of useless calculation is produced inevitably and noise could also be introduced, which inevitably reduces training efficiency and affects the final performance. For example, as shown in Figure \ref{intro}, a few red box patches contain the object, and most green box patches are background. The calculation between background patches is not useful for classification. From this point of view, \textbf{the processing of input images in ViT is not very suitable for fine-grained images}. 2) As a unique characteristic, ViT makes use of the predefined class token to predict. In a standard ViT model, class tokens are processed as same as each image patch in MSA module of all layers, and only the one from the last layer is used for classification. From one point of view, the class token used for classification is obtained based on all image patches under the self-attention manner, which may be unfavorable for it to further concentrate on critical subtle regions that are important for recognizing fine-grained categories. From the other point of view, according to our experiments, class tokens from different layers could extract features focusing on different information, and they are also complementary to each other. However, in FGVC, it will be much better if a model is able to leverage comprehensive features including both overall concepts and detailed information. Therefore, only taking the final class token is not enough to make full use of the feature extraction ability of ViT. As a result, \textbf{the standard classification manner of ViT with only class token in the final layer should be improved to learn and integrate more comprehensive information from fine-grained images for final prediction}.

Inspired by the above analyses, we propose a novel Vision Transformer based Fine-grained Object Discriminator (ViT-FOD) for fine-grained image classification tasks. Besides the ViT backbone, it further contains three novel modules, i.e., Complementary Tokens Integration (CTI), Attention Patch Combination (APC), and Critical Regions Filter (CRF). Specifically, CTI classifies objects based on class tokens from multiple layers instead of only the last one to integrate complementary information captured from different layers. APC destructs two images into patches and pieces the informative ones among them together to generate a new image. In this way, it reduces the influence of background in an input image by replacing corresponding regions with informative parts from another image. To some extent, APC can be treated as a data augmentation method that is more suitable for transformers because transformers are not as sensitive to objects' global structures as CNNs. CRF, similar to a localization module at a low computational cost, emphasizes tokens corresponding to discriminative regions to generate a new class token. According to experiments, ViT-FOD achieves the best classification accuracy up to now on several widely used fine-grained images datasets.

The main contributions of this paper are summarized below:
\begin{itemize}[topsep=0pt, noitemsep]
  \item We analyze the limitations of directly applying ViT to FGVC, and propose a novel framework, which can be trained end-to-end efficiently with only image labels.
  \item We propose a novel attentive patch combination module, which can be regarded as a novel data augmentation method more suitable for ViT. It optimizes the training efficiency of ViT applied to FGVC.
  \item We propose a complementary tokens integration module to integrate class tokens from different layers for complementary fine-grained feature extraction. To the best of our knowledge, this is the first attemp to explore and integrate different layers of ViT in FGVC tasks.
  \item We conduct extensive experiments on three widely used fine-grained image datasets and also make comprehensive analysis. The results demonstrate the proposed method is able to achieve competitive performance.
\end{itemize}

\section{Related Work}
As mentioned above, location-based methods are the mainstream for FGVC. Early works \cite{human-anno-1, human-anno-2} localize discriminative regions with the assistance of bounding boxes and part annotations. However, these methods are impractical due to the heavy involvement of manual annotations. Thereafter, some works locate the critical region in a weakly-supervised manner with only image labels.
Typical examples include RA-CNN\cite{racnn}, MA-CNN\cite{macnn}, MGE-CNN\cite{mge}, ELoPE\cite{elope} as well as DP-Net\cite{dpnet}. Although achieving passable results, these methods mainly focus on locating discriminative regions without considering how to integrate them into a unified concept. Recently, some studies try to establish spatial connections among localized regions. For instance, Stacked-LSTM\cite{stack-lstm} builds a bi-directional LSTM network to fuse and encode partial information of complementary parts into a comprehensive feature. Although it has achieved competitive results, its complicated structure has brought a significant increase in complexity in both time and space. Lately, the proposal of self-attention learning provides a new solution. For example, DNL\cite{dnl} splits attention into whitened pairwise term and unary term, and uses a disentangled non-local block which decouples the two terms to facilitate learning. GC-Net\cite{gcnet} simplifies the non-local network and presents a global context modeling framework.

\begin{figure*}
  \centering
  \includegraphics[width=0.9\textwidth]{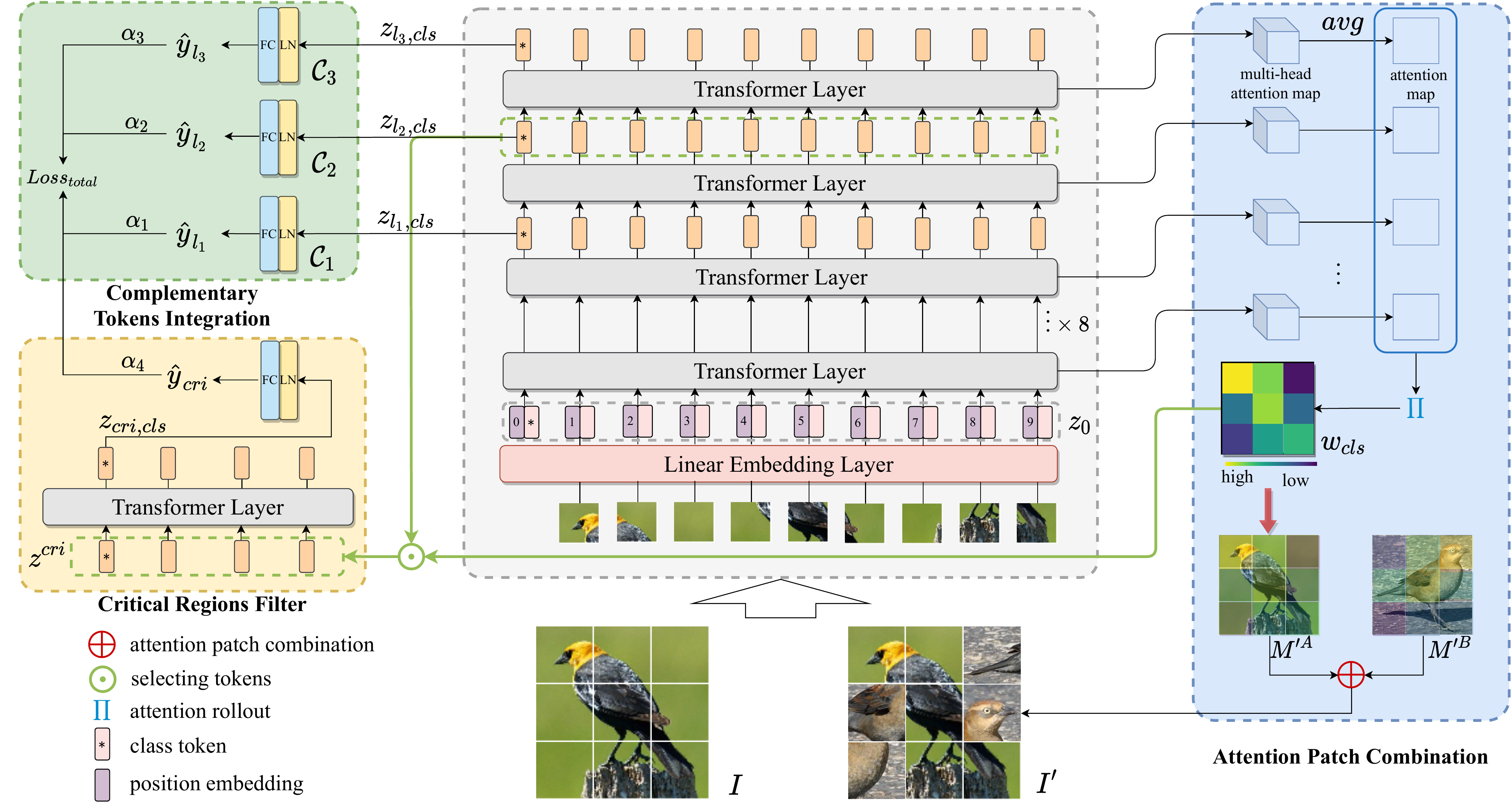}
  \caption{The overall framework of ViT-FOD.}
  \label{framework}
\end{figure*}

More recently, Transformer has been introduced into computer vision, i.e., Vision Transformer. Thereafter, it has been adopted in many fields and made significant achievements. Generally, ViT\cite{vit} first splits an image into a sequence of flattened patches and maps them into tokens, and then continuously calculates the relationship between tokens to obtain the description of the image. Because the image splitting and correlation calculation in ViT are suitable for fine-grained image feature learning naturally, some methods directly apply it to fine-grained datasets and have achieved satisfying results comparable with that of most CNN-based ones. For example, TransFG\cite{transfg} proposes the first ViT-based fine-grained classification method, in which a part selection module is designed to select discriminative tokens.
Following that, several methods\cite{ffvt, rams, tpskg} also apply ViT to FGVC. Thereinto, TPSKG\cite{tpskg} proposes a peak suppression module which penalizes the attention to the most discriminative part and a knowledge guidance module to obtain the knowledge response coefficients. RAMS-Trans\cite{rams} learns discriminative region attention in a multi-scale way. FFVT\cite{ffvt} aggregates the important tokens from each transformer layer to compensate for local, low-level and middle-level information.

Although these ViT-based methods have achieved promising performance, most of those methods simply inherit the design experience of CNN-based methods like peak suppression and discriminative region cropping, but lack consideration of the characteristics of the transformer itself. In contrast, in this paper, we discuss the limitations of ViT when directly applied to fine-grained datasets and propose a new ViT-based framework for fine-grained object discrimination.

\section{Method}

The overall framework of ViT-FOD is shown in Figure \ref {framework}. As mentioned above, besides the ViT backbone, ViT-FOD has three new modules, i.e, Complementary Tokens Integration (CTI), Attention Patch Combination (APC), and Critical Regions Filter (CRF). Thereinto, CTI integrates class tokens from different layers to obtain complementary information. APC selects informative regions to form new images. In addition, tokens with high attention are sent to the CRF module to generate a new class token with critical region emphasized. The details of those modules are described in the following sections.

\subsection{Complementary Tokens Integration}

Following ViT\cite{vit}, an image $I$ is first divided into $H\times W$ patches $x^i\in\mathbb{R}^{P\times P\times C},i\in\{1,...,N\}$, $P$ is the size of each patch, $C$ is the channel number of the image (usually 3), $N=H\times W$ is the number of patches. A linear embedding layer is applied to map each patch to a token. Besides, a learnable class token $x_{cls}$ is introduced for classification and position embedding is added to retain spatial information. Therefore, the input to the first Transformer Layer is as follows,
\begin{equation}
  z_0=[x_{cls}; x^1\boldsymbol{\mathrm{E}}; x^2\boldsymbol{\mathrm{E}};...; x^N\boldsymbol{\mathrm{E}}]+\boldsymbol{\mathrm{E}}_{pos},
\end{equation}
where $\boldsymbol{\mathrm{E}}\in \mathbb{R}^{(P^2\cdot C)\times D}$denotes the patch embedding projection, $D$ is the dimension of tokens, and $\boldsymbol{\mathrm{E}}_{pos}\in \mathbb{R}^{(N+1)\times D} $ is the position embedding. Suppose there are $L$ transformer layers. Each layer consists of a multi-head self-attention (MSA) block as well as a multi-layer perception (MLP) block. Both blocks have a Layer Normalization (LN) operation. And the output of each layer is,
\begin{equation}
  z'_l=\mathrm{MSA}(\mathrm{LN}(z_{l-1}))+z_{l-1}, \qquad l=1,...,L,
\end{equation}
\begin{equation}
  z_l=\mathrm{MLP}(\mathrm{LN}(z'_{l}))+z'_{l}, \qquad l=1,...,L.
\end{equation}

In the standard ViT model, the class token of the last layer is fed into a classifier to generate the labels, i.e., $\hat y=\mathcal C_L(z_{L,cls}) $, where $\hat y$ is the predicted label vector, $z_{L,cls}$ is the class token of the $L$-th layer and $\mathcal C_L$ represents the classifier.

Apparently, the above scheme ignores the class tokens learned in previous layers. However, they are also discriminative and could contain some information that the final class token loses. It means different layers could be complementary to each other, which is also verified in our experiments.
Inspired by this, we propose to make use of $k$ layers ($l_i,i\in\{1,...,k\}$) instead of only the last one to obtain more comprehensive fine-grained information. That is to say, the class token of each selected layer is sent to a classifier to generate a predicted label vector as follows,
\begin{equation}
  \hat{y}_{l_i}=\mathcal{C}_i(z_{l_i,cls}),\qquad i=1,...k.
\end{equation}
where $\hat y_{l_i}$ is the predicted label of the $l_i$-th layer, and $\mathcal{C}_i$ is the classifier for the $l_i$-th layer. In this way, we obtain multiple predictions of the input images at different layers. To make final prediction, we integrate the results of different layers. More specifically, the final decision is weighted by all predictions and the loss is
\vspace{-2mm}
\begin{equation}
  Loss_{CTI}=\sum^k_{i=1}{\alpha_i \mathrm{CE}(y, \hat{y}_{l_i})},
\end{equation}
where $\mathrm{CE}$ is the standard cross-entropy loss, $y$ represents the ground-truth label, and $\alpha_i$ is a weight parameter.

\subsection{Attention Patch Combination}

Transformer calculates the relationship between each pair of patches using multi-head self-attention. However, the calculation between the patches of background is unnecessary; therefore, there is much redundant calculation. To reduce it, APC module is designed to select informative patches from two images and combine them as a new input.

Given the input representation $z$ (for simplicity, the subscript is omitted), the self-attention is performed as follows. First, the query, key and value matrices $Q$, $K$ and $V$ are computed via linear projections,
\begin{equation}
  Q=zW^Q,K=zW^K,V=zW^V,
\end{equation}
where $W^Q$, $W^K$ and $W^V$ are weight matrices. Thereafter, the attention is gotten with the following formula,
\begin{equation}
  \mathrm {Attention} (Q,K,V)=\mathrm{softmax} (\frac{QK^T}{\sqrt{D}})V.
\end{equation}

Let $\mathrm {A}_l\in \mathbb{R}^{(N+1)\times(N+1)}$ is the attention map of the $l$-th layer. For multi-head self-attention, the attention map is $\mathrm{A}_l\in \mathbb R^{H_{head}\times(N+1)\times(N+1)}$ where $H_{head}$ is the number of heads. We add an identity matrix $E$ to the attention and average them to get the attention weight of each layer,
\begin{equation}
  W_{l}=\frac{\frac{1}{H_{head}}\sum_{i=1}^{H_{head}}\mathrm{A}_l+E}{2},
\end{equation}
where $W_{l}\in\mathbb{R}^{(N+1)\times(N+1)},l=1,...,L$, $W_{l,i,j}$ denotes the attention of the $i$-th token to the $j$-th token in the $l$-th layer. In order to get the ultimate weight map, we adopt the attention rollout algorithm\cite{rollout} which recursively applys a matrix multiplication to the attention weights of all the layers,
\begin{equation}
  W=\prod^L_{i=1}W_l,\qquad l=1,...,L.
\label{apc-w}
\end{equation}

\begin{figure}
  \centering
  \includegraphics[width=.43\textwidth]{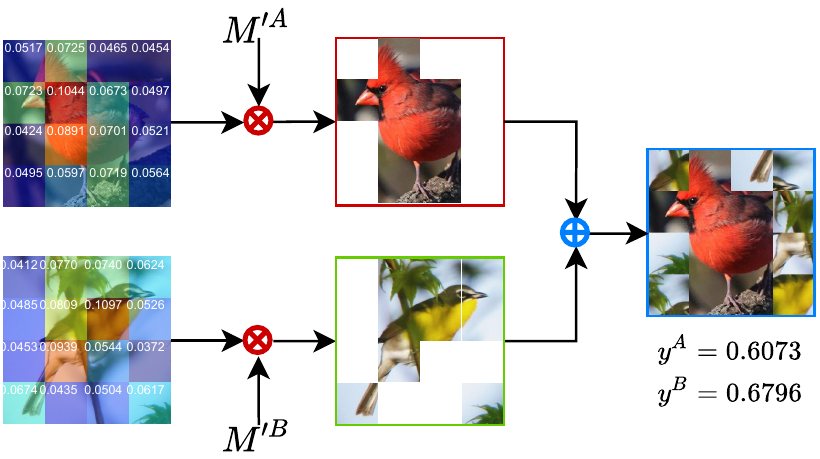}
  \caption{An example of Attention Patch Combination.}
  \label{apc}
\end{figure}

The APC aims to combine important patches of two images according to the weight map to eliminate redundant calculation. Moreover, APC can be treated as a data augmented method to improve the model's generalization ability. Specifically, after obtaining the $w_{cls}\in\mathbb{R}^{N}$ (the class token's attention to other tokens) from $W$, we reshape it into 2D and then pool it to $p\times p$ to get $w'_{cls}\in\mathbb{R}^{p\times p}$,
\begin{equation}\label{wclsp}
    w'_{cls}=\mathrm{Pool}_p(\mathrm{Reshape}_{2D}(w_{cls})).
\end{equation}

According to the weight map, we can get the corresponding serial number $idx_{cls}$ in descending order. For two images $I_A$ and $I_B$, we generate masks $M^A$ and $M^B$ in the following way,
\begin{equation}
  M_{i,j}=
  \begin{cases}
    1 & idx_{i,j}\le\frac{p^2}{2} \\
    0 & idx_{i,j}>\frac{p^2}{2}
  \end{cases}.
\end{equation}
In other words, if the weight of one patch belongs to the first half, it is set to 1, and vice versa. Then we generate a new image and its label with the following operations,
\begin{equation}
  I'=I_A\otimes M'^A\oplus I_B\otimes M'^B,
\end{equation}
\begin{equation}
  y'= (\sum w'^A_{cls}\otimes M^A)\cdot y^A +  (\sum w'^B_{cls}\otimes M^B)\cdot y^B,
\end{equation}
where $\otimes$ means element-wise multiplication, $\oplus$ means to fill the former's 0-value patches with the latter's 1-value patches in descending order. As illustrated in Figure \ref{apc}, the two masks are zoomed into the same size as the original image (written as $M'^A$ and $M'^B$), and then multiplied by the original image. After that, the patches in $I_B$ are filled into $I_A$ according to weight order. As for the label, we compute the corresponding weights by summing the weights of patches.


\subsection{Critical Regions Filter}
Cropping out discriminative regions and retraining model is an effective idea\cite{racnn,nts,dpnet} to emphasize subtle information in fine-grained images, which is also adopted by recently ViT-based RAMS-Trans\cite{rams}. However, this scheme may increase calculation cost significantly. Moreover, there is an apparent restriction problem of rectangle cropping. To address the problem, we propose a simple but effective Critical Regions Filter module to select the tokens of discriminative regions and generate an extra class token to collect information from the selected tokens. Compared with the cropping scheme, CRF allows a more flexible selection of patches rather than constrained to rectangular cropping.

Specifically, in order to focus on discriminative regions, we define a threshold $\eta (0<\eta\le1)$ to control the number of tokens selected, i.e., there are $\eta N$ tokens selected. Supposing the tokens are in descending order according to the weights in $w_{cls}$ and the weight of the $\eta N$-th token is $\bar{w}_{cls}$, we can get the mask of the selected tokens with the following operation, $M^{cri}\in\mathbb{R}^{N}$,
\begin{equation}
  M^{cri}_i=
  \begin{cases}
    1 & w_{cls,i}\ge \bar{w}_{cls} \\
    0 & w_{cls,i}< \bar{w}_{cls}
  \end{cases}.
\end{equation}

Finally, the selected tokens and the class token are concatenated as the input to the Transformer Layer in CRF, i.e.,
\begin{equation}
  z^{cri}=\mathrm{concat}(z_{L-1,cls};M^{cri}\odot(z_{L-1}\ominus z_{L-1,cls})),
\end{equation}
where $\odot$ represents the multiplication of the corresponding position and $\ominus$ is the operation of element removing. Thereafter, $z^{cri}$ is fed into an extra ViT layer and the class token in the output is further sent to a classifier to get a label prediction, i.e., $\hat{y}_{cri}$.
\begin{table}[t]
  \centering
  \caption{Statistics of three datasets.}
  \begin{tabular}{cccc}
    \toprule
    Datasets      & Class Number & Training & Testing \\
    \midrule
    CUB-200-2011\cite{cub}  & 200          & 5,994     & 5,794    \\
    Stanford Dogs\cite{dog} & 120          & 12,000    & 8,580    \\
    NABirds\cite{nabirds}       & 555          & 23,929    & 24,633   \\
    \bottomrule
  \end{tabular}%
  \label{tab:datasets}%
\end{table}%
\subsection{Training and Inference}
As mentioned above, APC module can be treated as a data augmentation method. In the training phase, the generated images (with their corresponding labels) and original ones are all used as input. In addition, the total loss is composed of CTI loss and CRF loss,
\begin{align}
  Loss_{total} & =Loss_{CTI}+Loss_{CRF}\notag  \\                                                                  & =\sum_{i=1}^k{\alpha_i\mathrm{CE}(y,\hat y_{l_i})} + \alpha_{k+1}\mathrm{CE}(y,\hat y_{cri}).
\end{align}

During inference, there is no APC, and the classification result is composed of the weighted output of predictions,
\begin{equation}
    \hat{y} =\sum^k_{i=1}\beta_i \hat y_{l_i}+\beta_{k+1} \hat y_{cri},
\end{equation}
where $\beta_i$ is a weight parameter.
\section{Experiments}
\subsection{Experimental Settings}
\label{sec4-1}

\textbf{Datasets.} We evaluated our proposed ViT-FOD on three widely used fine-grained datasets, namely, CUB-200-2011\cite{cub}, Stanford Dogs\cite{dog}, and NABirds\cite{nabirds}. The details are summarized in Table \ref{tab:datasets}.

\begin{table}[t]
  \centering
  \caption{Accuracy of different methods on CUB-200-2011.}
  \begin{tabular}{ccc}
    \toprule
    Method                            & Backbone      & Accuracy(\%)  \\
    \midrule
    B-CNN\cite{bcnn}                   & VGG-16,VGG-19 & 85.1          \\
    RA-CNN\cite{racnn}                 & VGG-19        & 85.3          \\
    MA-CNN\cite{macnn}                 & VGG-19        & 86.5          \\
    PA-CNN\cite{pacnn}                 & VGG-19        & 87.8          \\
    NTS-Net\cite{nts}                     & ResNet-50     & 87.5          \\
    Cross-X\cite{cross-x}             & ResNet-50     & 87.7          \\
    DCL\cite{dcl}                     & ResNet-50     & 87.8          \\
    ACNet\cite{acnet}                 & ResNet-50     & 88.1          \\
    AP-CNN\cite{apcnn}                 & ResNet-50     & 88.4          \\
    S3N\cite{s3n}                     & ResNet-50     & 88.5          \\
    SPS\cite{sps}                     & ResNet-50     & 88.7          \\
    DP-Net\cite{dpnet}                & ResNet-50     & 89.3          \\
    PMG\cite{pmg}                     & ResNet-50     & 89.6          \\
    CIN\cite{cin}                     & ResNet-101    & 88.1          \\
    ELoPE\cite{elope}                 & ResNet-101    & 88.5          \\
    MGE-CNN\cite{mge}                 & ResNet-101    & 89.4          \\
    CAL\cite{cal}                     & ResNet-101    & 90.6          \\
    FDL\cite{fdl}                     & DenseNet-161   & 89.1         \\
    API-Net\cite{api-net}             & DenseNet-161   & 90.0          \\
    Stacked-LSTM\cite{stack-lstm}             & GoogLeNet      & 90.4          \\
    \midrule
    ViT\cite{vit}                     & ViT-B\_16     & 89.8          \\
    RAMS-Trans\cite{rams}                   & ViT-B\_16     & 91.3          \\
    TPSKG\cite{tpskg}                 & ViT-B\_16     & 91.3          \\
    FFVT\cite{ffvt}                   & ViT-B\_16     & 91.6          \\
    TransFG\cite{transfg}             & ViT-B\_16     & 91.7          \\
    \midrule
    ViT-FOD                              & ViT-B\_16     & \textbf{91.8} \\
    ViT-FOD$^\dagger$                    & ViT-B\_16     & \textbf{91.9} \\
    \bottomrule
  \end{tabular}%
  \label{tab:cub}%
\end{table}%

\textbf{Implementation details.} During training, we scaled the short edge of the input to 512 while maintaining the length-width ratio, and then randomly cropped a region with the size of 448*448, and flipped it horizontally. In inference phase, random cropping was substituted with center cropping. ViT-B\_16\cite{vit} was used as backbone and initialized with pretained weights on ImageNet21k. In addition to the 12-th layer, the 10-th and 11-th layers were selected, i.e., $k=3$, $l_1=10$, $l_2=11$, and $l_3=12$. In CRF, the transformer layer was initialized to be the same as the $12$-th layer and the threshold $\eta$ was set to 0.2. The parameters were selected based on evaluations on training data. Specifically, for CUB-200-2011 and NABirds, $\alpha _i = 1$, as for Stanford Dogs, $\alpha_1 = \alpha_2 = 0.01$. For CUB-200-2011, $\beta_1 = \beta_2 = 0.2, \beta_3 = \beta_4 = 0.8$, and for Stanford Dogs and NABirds, $\beta _i = 0.5$. The batch size was 10. SGD was adopted as optimizer with momentum=0.9, weight decay= $5e^{-4}$. The  learning rate was initialized as $3e^{-4}$ except $3e^{-2}$ for the fully connected layer of the 10-th and 11-th layers on Stanford Dogs. Cosine annealing was adopted to train 60 epochs totally. Experiments were performed on NVIDIA GeForce RTX 2080 GPUs with Pytorch.



\begin{table}[t]
  \centering
  \caption{Accuracy of different methods on Stanford Dogs.}
  \begin{tabular}{ccc}
    \toprule
    {\qquad Method \qquad}               & Backbone    & Accuracy(\%)  \\
    \midrule
    RA-CNN\cite{racnn}     & VGG-19      & 87.3          \\
    DB-GCE\cite{db-gce}   & ResNet-50   & 87.7          \\
    SEF\cite{sef}         & ResNet-50   & 88.8          \\
    Cross-X\cite{cross-x} & ResNet-50   & 88.9          \\
    MAMC\cite{mamc}       & ResNet-101  & 85.2          \\
    API-Net\cite{api-net} & ResNet-101 & 90.3          \\
    MaxEnt\cite{maxent}   & DenseNet-161 & 83.6         \\
    FDL\cite{fdl}         & DenseNet-161 & 84.9         \\
    \midrule
    TransFG\cite{transfg} & ViT-B\_16   & 92.3          \\
    ViT\cite{vit}         & ViT-B\_16   & 92.4          \\
    RAMS-Trans\cite{rams}       & ViT-B\_16   & 92.4          \\
    TPSKG\cite{tpskg}     & ViT-B\_16   & 92.5          \\
    \midrule
    ViT-FOD               & ViT-B\_16   & \textbf{92.9} \\
    ViT-FOD$^\dagger$        & ViT-B\_16   & \textbf{93.0} \\
    \bottomrule
  \end{tabular}%
  \label{tab:dog}%
\end{table}%

\subsection{Comparison with State-of-the-Art.}

The results compared with other methods on CUB-200-2011 are shown in Table \ref{tab:cub}. Here, we divide all methods into three groups, i.e., convolution-based methods, transformer-based methods, and our proposed method. Thereinto, ViT-FOD$^\dagger$ represents ViT-FOD adopts the Sharpness-Aware Minimization (SAM) optimizer\cite{sam}. From Table \ref{tab:cub}, we have the following observations.
\begin{itemize}
\item The original ViT-B\_16 which is the backbone in transformer-based methods already achieves an accuracy of 89.8\%, and outperforms most of the convolution-based methods, demonstrating the superiority of ViT in dealing with fine-grained data.
\item It is worth noting that TransFG is the first work applying transformer to FGVC, which is also the best baseline in Table \ref{tab:cub}. However, it uses overlapping patches, which may lead to a high cost of computation.
\item ViT-FOD achieves the highest accuracy of 91.8\% without overlapping. Moreover, the accuracy of ViT-FOD can be further improved to 91.9\% when it adopts the Sharpness-Aware Minimization (SAM) optimizer, i.e., ViT-FOD$^\dagger$.

\end{itemize}


As mentioned previously, ViT-FOD is able to reduce the calculation by selecting informative patches instead of using overlapping ones. To demonstrate the efficiency of ViT-FOD, we made statistics on three models. The results are summarized in Table \ref{tab:complex}. We observe that, compared with ViT-B\_16, the flops of ViT-FOG increases a little while the flops of TransFG is 1.6 times as much as ViT-B\_16. The number of parameters of ViT-FOD increased by only 8\%. In a word, the test efficiency of ours is better than TransFG with a similar number of parameters.

\begin{table}[t]
  \centering
  \caption{Accuracy of different methods on NABirds.}
  \begin{tabular}{ccc}
    \toprule
    Method                      & Backbone    & Accuracy(\%)  \\
    \midrule
    Cross-X\cite{cross-x}       & SENet-50   & 86.4          \\
    PAIRS\cite{pairs}           & ResNet-50   & 87.9          \\
    SPS\cite{sps}               & ResNet-101  & 87.9         \\
    MGE-CNN\cite{mge}           & ResNet-101  & 88.6          \\
    CS-Parts\cite{cs-parts}     & Inception-v3   & 88.5          \\
    MaxEnt\cite{maxent}         & DenseNet-161 & 83.0        \\
    API-Net\cite{api-net}       & DenseNet-161 & 88.1          \\
    FixSENet-154\cite{fixsenet} & SENet-154   & 89.2          \\
    \midrule
    ViT\cite{vit}               & ViT-B\_16   & 89.1          \\
    TPSKG\cite{tpskg}           & ViT-B\_16   & 90.1          \\
    TransFG\cite{transfg}       & ViT-B\_16   & 90.8          \\
    \midrule
    ViT-FOD                       & ViT-B\_16   & \textbf{91.4} \\
    ViT-FOD$^\dagger$              & ViT-B\_16   & \textbf{91.5} \\
    \bottomrule
  \end{tabular}%
  \label{tab:nabirds}%
\end{table}%

The results on Stanford Dogs and NABirds are shown in Table \ref{tab:dog} and \ref{tab:nabirds}, respectively. Similar to the results on CUB-200-2011, transformer-based methods perform better than most convolution-based ones significantly, and ViT-FOD outperforms all baselines. Specifically, it outperforms the best baseline TPSKG 0.4\% on Stanford Dogs and TransFG 0.6\% on NABirds. Furthermore, when SAM is adopted as the optimizer, ViT-FOD gets a steady 0.1\% performance improvement.

\subsection{Ablation study}

\subsubsection{Contribution of each module}
\
\indent There are three new proposed modules in ViT-FOD, i.e., CTI, APC, and CRF.
To demonstrate their effectiveness, we conducted ablation studies on CUB-200-2011 and reported results in Table \ref{tab:components}. From the table, we can see that all three components are effective, improving ViT-B\_16\cite{vit} by 1.4\%, 0.9\%, and 1.4\%, respectively. Better performance can be obtained by combining these modules. Among them, the combination of CTI and CRF obtains less performance gain. The combinations of APC with CTI and CRF achieve accuracy of 91.7\% and 91.6\%, respectively. And when three components are used, the model gets the highest accuracy 91.8\%.

\subsubsection{Architecture of the CTI module}
\
\indent  The motivation of CTI is to integrate class tokens from different layers to obtain complementary information. We conducted experiments to demonstrate that different layers indeed contribute to the performance.

\begin{figure}
  \centering
  \includegraphics[width=.4\textwidth]{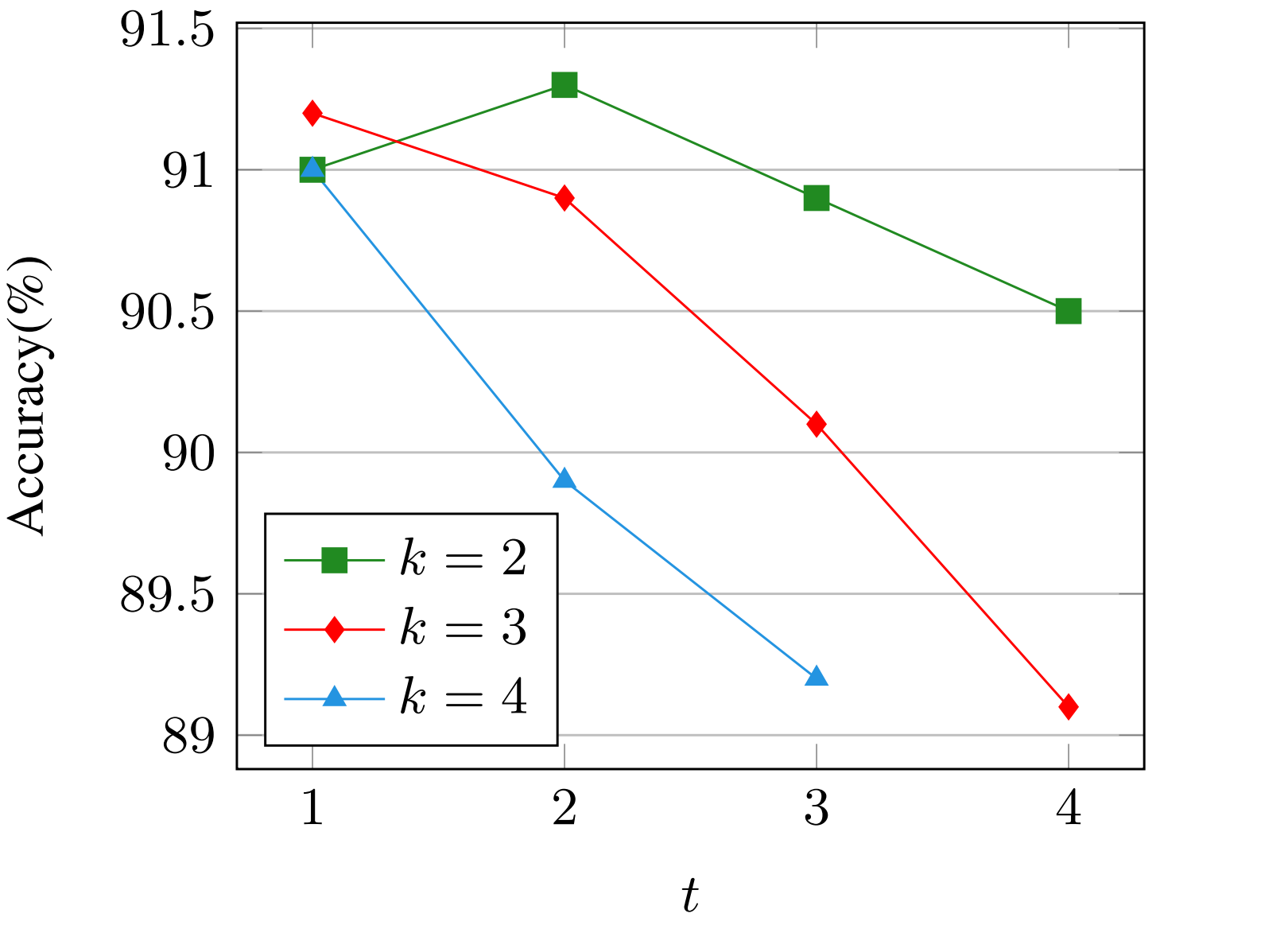}
  \vspace{-0.3cm}
  \caption{Accuracy of different settings of CTI on CUB-200-2011.}
  \label{ml}
\end{figure}

For ViT-B\_16, there are $2^{12}-1$ layer combination ways in total, which are impossible to verify one by one. For simplicity, we selected the $12$-th layer as a benchmark and selected other layers with a step size of $t$. For example, the $10$-th and $12$-th layers are selected when $t=2$. The experimental results of different layer combinations are shown in Figure \ref{ml}, where $k$ is the total number of selected layers. We can observe that the smaller the interval between layers, the better the performance tends to be. For example, when $k$ is set to 3, $t=1$ is a better choice than $t=2$ and $t=3$.
Besides, more layers do not lead to better performance. For example, when $t$ is set to 2, the performance of $k=4$ is 1\% worse than $k =3$. One of the main reasons could be that, if $t$ or $k$ is too large, the bottom layers will participate in training and inference. However, these layers have not had a clear attention map to the object and are susceptible to noise. At the same time, the training of top layers could also be influenced. Among all the possible choices, ViT-FOD is more robust when $k=3$, $t=1$.

\begin{table}[t]
  \centering
  \caption{Model complexity comparison.}
    \begin{tabular}{cccc}
    \toprule
    Model & ViT-B\_16 & TransFG & ViT-FOD \\
    \midrule
    Flops(G) & 67.14 & 107.56 & 68.25 \\
    Parameters(M) & 85.76 & 85.76 & 93.31 \\
    \bottomrule
    \end{tabular}%
  \label{tab:complex}%
\end{table}%

\begin{table}[t]
  \centering
  \caption{Performance of components on CUB-200-2011.}
  \begin{tabular}{ccccc}
    \toprule
    ViT-B\_16  & CTI         & CRF         & APC        & Accuracy(\%) \\
    \midrule
    \checkmark &            &            &            & 89.8         \\
    \checkmark & \checkmark &            &            & 91.2         \\
    \checkmark &            & \checkmark &            & 90.7         \\
    \checkmark &            &            & \checkmark & 91.2         \\
    \checkmark & \checkmark & \checkmark &            & 91.3         \\
    \checkmark & \checkmark &            & \checkmark & 91.6         \\
    \checkmark &            & \checkmark & \checkmark & 91.7         \\
    \checkmark & \checkmark & \checkmark & \checkmark & 91.8         \\
    \bottomrule
  \end{tabular}%
  \label{tab:components}%
\end{table}%

In addition, we also recorded the accuracy of different layers in Table \ref{tab:ml}. The best performance is obtained when three layers are combined due to the complementarity of attention, demonstrating that different layers provide complementary information.

\begin{figure}
  \centering
  \includegraphics[width=.42\textwidth]{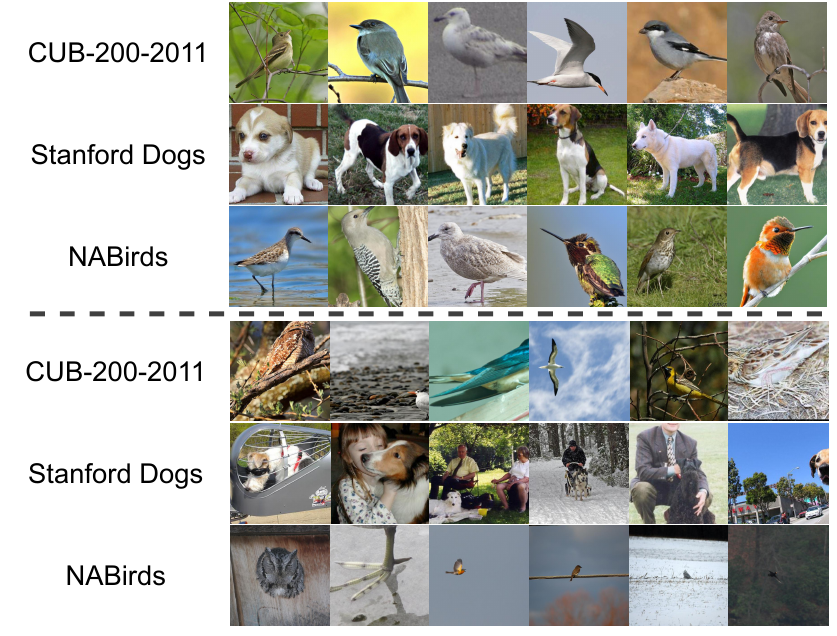}
  \caption{Predicted samples from three datasets. Images of the first three rows are classified correctly at the $10$th layer and incorrectly at the $12$th layer. The last three rows are the opposite.}
  \label{1012layer}
\end{figure}

Besides, we also visualized the correctly predicted images of the $10$-th and $12$-th layers, as shown in Figure \ref{1012layer}. We observe that the images classified correctly at the $10$-th layer are generally clear and the postures of the objects are relatively standard. In these cases, excessive attention to some regions may lead to misclassification. In contrast, images classified correctly at the $12$-th layer either have a complex scene or have a small even incomplete object. Therefore, these images need strong attention. This further demonstrates that class tokens from different layers in ViT are complementary to each other.

\begin{table}
  \centering
  \caption{Accuracy of different layers}
  \begin{tabular}{cccc}
    \toprule
    \multirow{2}[4]{*}{layer} & \multicolumn{3}{c}{Accuracy(\%)}                   \\
    \cline{2-4}            & CUB-200-2011                              & Stanford Dogs   & NABirds \\
    \hline
    10                     & 90.7                            & 53.6 & 90.6   \\
    11                     & 91.1                            & 87.8 & 91.0   \\
    12                     & 91.6                            & 92.7 & 91.0   \\
    $\sum$                 & 91.8                            & 92.9 & 91.4   \\
    \bottomrule
  \end{tabular}%
  \label{tab:ml}%
\end{table}%

\begin{table}[t]
  \centering
  \caption{Performance impact of $\eta$ on CUB-200-2011.}
  \begin{tabular}{cccccc}
    \toprule
    $\eta$       & 1    & 0.9  & 0.8  & 0.7  & 0.6  \\
    \midrule
    \ \ Accuracy(\%)\ \ \ & -    & 89.9 & 90.0 & 90.0 & 90.0 \\
    \midrule
    \toprule
    $\eta$       & 0.5  & 0.4  & 0.3  & 0.2  & 0.1  \\
    \midrule
    Accuracy(\%) & 90.4 & 90.3 & 90.4 & 90.7 & 90.4 \\
    \bottomrule
  \end{tabular}%
  \label{tab:eta}%
\end{table}%

\begin{figure*}
  \centering
  \includegraphics[width=1.0\textwidth]{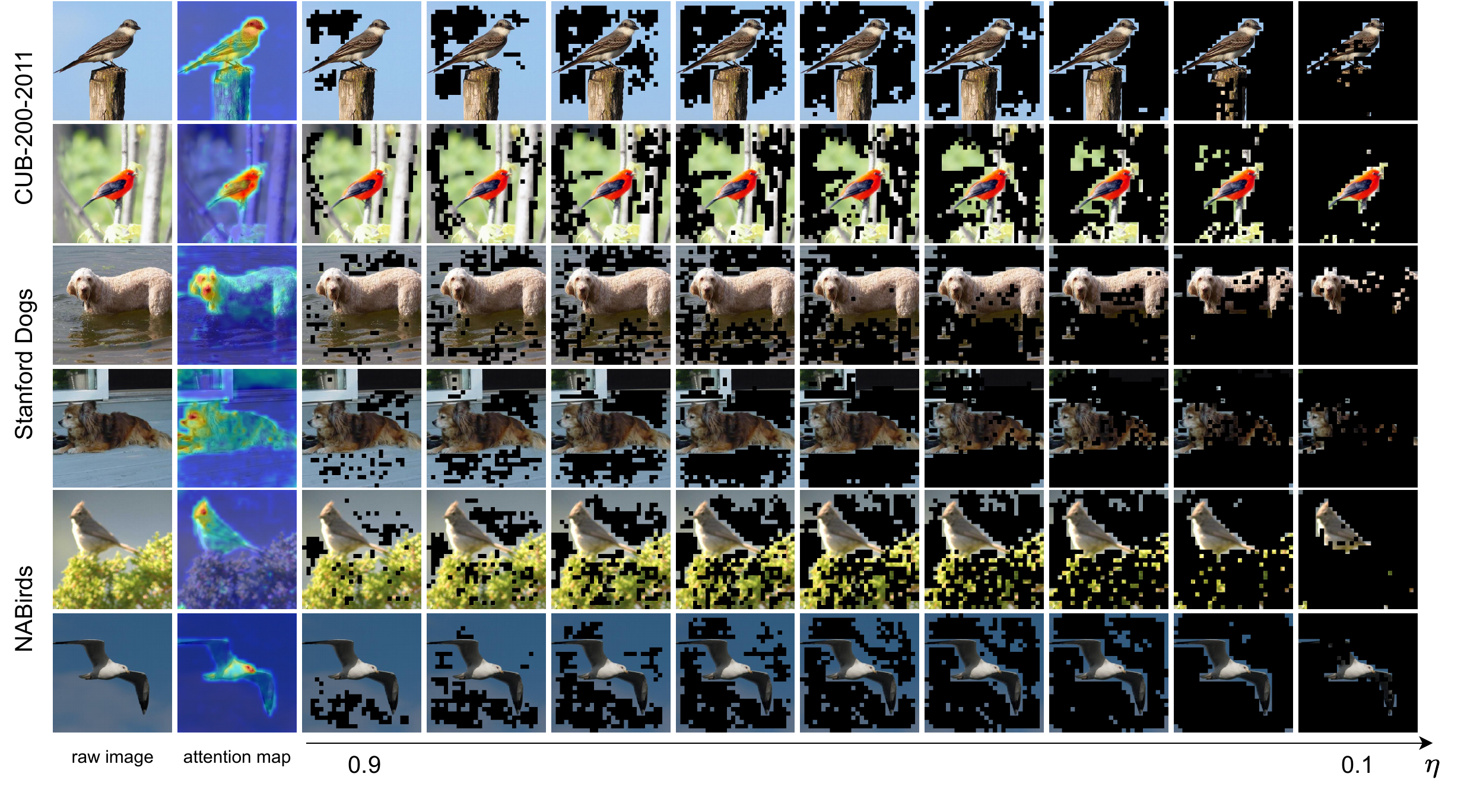}
  \caption{Visualization results of CRF with different $\eta$ on three datasets.}
  \label{lb}
\end{figure*}

\subsubsection{The threshold $\eta$ in CRF}
\
\indent In CRF module, $\eta$ determines the number of tokens sent to CRF. To evaluate its impact on performance, we conducted experiments on CUB-200-2011 by setting $\eta$ to different values.
The results are shown in Table \ref{tab:eta}. When $\eta$ is large, more tokens are sent to CRF; however, the impact on performance is small. As $\eta$ decreases, fewer tokens are sent to CRF, meaning that images are classified according to some important patches with fine-grained information. ViT-FOD achieves the best result when $\eta=0.2$.


To further demonstrate that CRF indeed plays a role in reducing meaningless tokens, we resize the mask generated by CRF and put it on the original image. Several samples are visualized in Figure \ref{lb}. We observe that most patches with high attention are objects. As $\eta$ decreases, patches with lower attention are deleted, most of which are background and useless for classification. The left patches with higher attention are the key ones of objects. To consider the patches with different attention, general methods \cite{rams} usually crop and zoom the largest connected component and use it for training again. In this way, the cropped image inevitably contains background because an object does not appear as a standard rectangle to a large extent. In contrast, ViT-FOD discards those background patches by adjusting parameter $\eta$. Moreover, unlike convolution methods, which require the input to be a 3D feature map, the Transformer Layer in CRF does not require the number of tokens to be fixed; therefore, it is more flexible in real applications.


\begin{table}[t]
  \centering
  \caption{Performance impact of $p$ in APC on CUB-200-2011.}
    \begin{tabular}{ccccccc}
    \toprule
    $p$     & 3     & 4     & 5     & 6     & 7  \\
    \midrule
    Accrracy(\%) & 91.70  & 91.80  & 91.73  & 91.73  & 91.75 \\
    \bottomrule
    \end{tabular}%
  \label{tab:p}%
\end{table}%


\vspace{-3mm}
\subsubsection{Patch number $p$ in APC}
\
\indent Parameter $p$ 
controls the number of parts that image is split in APC.
To investigate its impact on performance, we further conducted experiments on CUB-200-2011. The results are summarized in Table \ref{tab:p}. From Table \ref{tab:p}, we can observe that ViT-FOD obtains the best accuracy when $p=4$; however, it is not very sensitive to $p$. Intuitively, if $p$ is too small, each patch is large; thus the attention cannot well describe the information of every patch. If $p$ is big, each patch is small which implies finer attention. However, when $p$ is large, meaningful parts are possibly destroyed and it takes more processing time to perform patch swapping. Therefore, in our experiments, we adopted a compromise setting, i.e., $p=4$.

\subsubsection{Comparison with other mix-based methods}
\
\indent As aforementioned, APC module combines informative regions from two images to generate a new input that works like a data augmentation method. In order to demonstrate its superiority over other data augmentation strategies, we compare APC with widely-adopted cutout\cite{cutout}, cutmix\cite{cutmix} and mixup\cite{mixup}. Experimental results are summarized in Table \ref{tab:mix}.
The accuracy of cutout is even lower than that of baseline. Because cutout randomly erases regions, it may not be suitable for fine-grained images, resulting in performance degradation.
Although both mixup and cutmix improve the baseline, the performance gain of APC is more significant, demonstrating that APC is more suitable for ViT. In fact, although both APC and cutmix aim to augment inputs by integrating different images, their motivation and realization are different. Firstly, cutmix only exchanges a whole region, which is limited by the integrity of CNNs. By contrast, ViT is less dependent on global location information. Therefore, it can exchange several scattered patches. Secondly, in cutmix, the object may be completely covered by the background from another image; however, APC avoids this by replacing background patches with informative ones. Thirdly, due to the diversity of object sizes and randomness of cropping, it is unreasonable for cutmix to use the area as the label. In contrast, APC calculates labels of generated images, which is a more reliable way.
\begin{table}[t]
  \centering
  \caption{Comparison with other mix-based methods.}
  \begin{tabular}{cccccc}
    \toprule
    method       & {ViT-B\_16} & {cutout} & {mixup} & {cutmix}  & {APC} \\
    \midrule
    Accuracy(\%) & 89.8  &  89.4   & 90.1    & 90.3    & 91.2  \\
    \bottomrule
  \end{tabular}%
  \label{tab:mix}%
  \vspace{-0.4cm}
\end{table}%

\section{Conclusion}

In this work, we analyze the shortcomings of Vision Transformer directly applied to FGVC and propose a novel ViT-based framework for FGVC tasks. In this framework, three new modules are designed. Thereinto, CTI module is able to make full use of discriminative and complementary class tokens at multiple layers. APC replaces the irrelevant background with informative patches from another image. In this way, it can not only augment the inputs, but also reduce redundancy computing. In addition, CRF module emphasizes those tokens of discriminative regions, making the model focus more on fine-grained feature. Moreover, it also improves training efficiency. Comprehensive experiments are conducted on three fine-grained image datasets and the results demonstrate ViT-FOD achieves SOTA performance.

\balance
\bibliographystyle{ACM-Reference-Format}
\bibliography{sample-base}


\end{document}